\documentclass[letter, 10pt, conference]{ieeeconf}      
\IEEEoverridecommandlockouts %
\usepackage{times}
\usepackage{graphicx}\graphicspath{{img/}}
\usepackage{epstopdf}
\usepackage{algorithm}
\usepackage{algpseudocode}
\usepackage{multirow} 
\usepackage{multicol}
\usepackage{array}
\usepackage{mdwmath}
\usepackage{mdwtab}
\usepackage{rotating}
\usepackage[list=off]{caption}
\usepackage{amssymb}
\usepackage{verbatim}
\usepackage{overpic}
\usepackage{amsmath}
\usepackage{subcaption}
\usepackage[utf8]{inputenc}
\usepackage[T1]{fontenc}
\usepackage[export]{adjustbox}
\usepackage[normalem]{ulem}
\usepackage{siunitx} 
\usepackage{chngpage}
\usepackage{xcolor}

\usepackage{xspace}

\providecommand{\HCI    }{human computer interfaces}

\usepackage[acronym]{glossaries}%
\glsdisablehyper 
\makeglossaries%

\newacronym{AI}{AI}    {artificial intelligence}%
\newacronym{RL}{RL}    {reinforcement learning}\providecommand{\RL}{\gls{RL}}%
\newacronym{HRI}{HRI}  {human robot interaction}%
\newacronym{LfD}{LfD}  {learning from demonstration}\providecommand{\LfD}{\gls{LfD}}%
\newacronym{RLfD}{RLfD}{reinforcement learning from demonstration}\providecommand{\RLfD}{\gls{RLfD}}%
\newacronym{MT}{MT}    {machine teaching}\providecommand{\MT}{\gls{MT}}%
\newacronym{NMSE}{NMSE}{normalised mean squared error}%
\newacronym{OLS} {OLS} {ordinary least squares}%
\newacronym{LiP}{LiP}  {linear-in-the-parameters}%
\newacronym{SVM}{SVM}  {support vector machine}%
\newacronym{LQR}{LQR}  {linear quadratic regulator}%
\newacronym{LSPI}{LSPI}{least squares policy iteration}\providecommand{\LSPI}{\gls{LSPI}}%
\newacronym{AR}{AR}    {augmented reality}\providecommand{\AR}{\gls{AR}}%
\newacronym{ADE}{ADE}  {absolute demonstration error}%
\newacronym{ARMSE}{ARMSE}{average root mean square error}\providecommand{\ARMSE}{\gls{ARMSE}}
\newacronym{ATC}{ATR}    {average total reward}%
\newacronym{HCI}{HCI}    {human-computer interaction}\providecommand{\HCI}{\gls{HCI}}%
\usepackage{amssymb}
\usepackage{mathtools}
\usepackage{amsmath}
\usepackage{gensymb}
\usepackage{algorithm}
\usepackage{algpseudocode}
\usepackage{amsfonts}
\usepackage{bm}

\mathchardef\mhyphen="2D   

\providecommand{\R}     {\mathbb{R}}          
\providecommand{\T}     {\top}                
\providecommand{\U}     {\mathcal{U}}         
\providecommand{\I}     {\bm{I}}          

\providecommand{\estimated} [1]{\tilde{#1}}

\providecommand{\optimal}   [1]{\bar{#1}}

\providecommand{\nd}      {n}                              
\providecommand{\Nd}      {\MakeUppercase{\nd}}            

\providecommand{\tk}     {k}                        
\providecommand{\nx}     {p}                  

\providecommand{\bA}     {\bm{A}}         
\renewcommand  {\r}      {r}              
\providecommand{\br}     {\bm{\r}}        

\providecommand{\bB}     {\bm{B}}         

\providecommand{\u}      {u}                  

\providecommand{\bu}     {\bm{\u}}        
\providecommand{\nu}     {q}\renewcommand{\nu}{q}        
\providecommand{\dimu}   {\mathcal{\MakeUppercase{\nu}}} 

\providecommand{\j}{j}\renewcommand{\j}      {j}                  

\providecommand{\A}     {\mathcal{A}} 
\providecommand{\D}     {\bm{\mathcal{D}}} 

\providecommand{\desired}[1]{#1^{*}}%
\providecommand{\argmin}{\mathop{\rm arg~min}\limits}
\providecommand{\argmax}{\mathop{\rm arg~max}\limits}
\providecommand{\bphi}  {\bm{\phi}}
\providecommand{\bpsi}  {\bm{\psi}}
\providecommand{\bPsi}  {\bm{\Psi}}

\providecommand{\triskf}  {\mathcal{\rho}}    

\providecommand{\terrorf}  {\mathcal{\varepsilon}}    
\providecommand{\truestate}{\bm{x}^{*}}           
\providecommand{\state}{\bm{x}}           
\providecommand{\nextstate}{\state^\prime} 
\providecommand{\dimx}   {\mathcal{\MakeUppercase{\nx}}} 
\providecommand{\learntstate}{\estimated{\state}}  

\providecommand{\action}{\bu}  
\providecommand{\dimu}{\mathcal{Q}} 

\providecommand{\model}{\bm{\theta}}
\providecommand{\target}{\bf{\optimal{\model}}}  
\providecommand{\learnt}{\bf{\estimated{\model}}}  
\providecommand{\dimphi}{\mathcal{S}} 
\providecommand{\dimpsi}{\mathcal{T}} 
\providecommand{\policy}    {\bm{\pi}}   
  
\providecommand{\featurepi}{\bphi}
\providecommand{\dimfeaturepi}{\dimphi}
\providecommand{\dataset}{\D}
\providecommand{\bestdataset}{\optimal{\D}}
\providecommand{\Nd}{\mathcal{N}} 
\providecommand{\numD} {\Nd} 

\providecommand{\searchspaceD}{\bm{\mathfrak{D}}} 

\providecommand{\valuefunction}{Q^{\policy}} 
\providecommand{\cost}{\j}
\providecommand{\costs}{\bm{\cost}}%
\providecommand{\humancost}{\estimated{\jmath}} 

\providecommand{\humancosts}{\estimated{\bm{\jmath}}} 

\providecommand{\truecost}{\optimal{\jmath}} 
\providecommand{\truecosts}{\optimal{\bm{\jmath}}} 
\providecommand{\modelq}{\bm{\omega}}
\providecommand{\truemodelq}{\optimal{\modelq}}%
\providecommand{\Featureq}{\bPsi}
\providecommand{\featureq}{\bpsi}
\providecommand{\dimfeatureq}{\dimpsi}
 
\providecommand{\nextFeatureq}{\Featureq^\prime}%

\providecommand{\bQ}{\bm{Q}}%
\providecommand{\bR}{\bm{R}}%

\providecommand{\Tk}{\MakeUppercase{\tk}}

\providecommand{\Eade}  {\mathnormal{E_{\acrshort{ADE}}}}
\providecommand{\armse}  {\mathnormal{E_{ARMSE}}} 

\providecommand{\eltwo}  {\mathnormal{\ell_2}}
\providecommand{\atc}  {\mathnormal{E_{\acrshort{ATC}}}}

\providecommand{\statei}    {\r_1}
\providecommand{\stateii}   {\r_2}
\providecommand{\ui}      {\delta\statei}
\providecommand{\uii}     {\delta\stateii}

\providecommand{\bPsi}  {\bm{\Psi}}               %
\providecommand{\rotation}  {\bm{T}(\alpha)}

\DeclareUnicodeCharacter{0308}{??}

\FPset{\trialsi}{500} 

\FPset{\exptiNd}{8}
\usepackage{xspace}

\providecommand{\ie}{\textit{i.e.,}~} %
\providecommand*{\sref}[1]{\S\ref{s:#1}}            

\providecommand{\figurename}{Fig.}
\providecommand*{\fref}[1]{\figurename~\ref{f:#1}}  
\providecommand*{\eref}[1]{(\ref{e:#1})}            

\let\labelindent\relax 
\usepackage[inline]{enumitem} 
\setlist{nolistsep}
\providecommand{\il}[1]{\begin{enumerate*}[label=(\roman*)]#1\end{enumerate*}} 
\providecommand{\cl}[1]{\begin{enumerate*}[label=(\alph*)]#1\end{enumerate*}}  
\usepackage{caption}
\providecommand{\phaseref}[1]{P\ref{phase:#1}}%
\providecommand{\skillref}[1]{S\ref{skill:#1}}%

\usepackage[normalem]{ulem}                                        
\usepackage{marginnote}
\usepackage[textwidth=\marginparwidth,colorinlistoftodos]{todonotes}

\colorlet{jb}{red}
\colorlet{mh}{red}
 
\providecommand  {\colorsout}[1]{\bgroup\markoverwith{\textcolor{#1}{\rule[0.5ex]{2pt}{0.4pt}}}\ULon} 

\colorlet{ma}{blue}

\makeatletter\newcommand{\manuallabel}[2]{\def\@currentlabel{#2}\label{#1}}\makeatother

\usepackage[
backend=bibtex,
style=ieee,
url=false,
doi=false,
isbn=false,
doi=true,
eprint=false,
maxcitenames=1,mincitenames=1
]{biblatex}
\renewcommand{\cite}[1]{~\autocite{#1}}

\usepackage[backend=bibtex]{biblatex}
\usepackage{graphicx}
\usepackage{overpic}
\usepackage{gensymb}
\usepackage[acronym]{glossaries}
\usepackage{pgfplots}
\usepackage{amsmath}
\usepackage{xcolor}
\definecolor{custompink}{HTML}{FF69B4}
\usepackage{enumitem}

\usepackage{filecontents}
\usepgfplotslibrary{fillbetween}
\usetikzlibrary{arrows.meta}

\usepackage{pgfplots}
\usepgfplotslibrary{statistics}
\pgfplotsset{compat=1.18}

\title{\LARGE \bf
Training People to Reward Robots}

\author{Endong Sun$^{1}$, Yuqing Zhu and Matthew Howard
\thanks{$^{1}$Endong Sun ({\tt\small endong.sun@kcl.ac.uk}), Yuqing Zhu and Matthew Howard are with the Centre for Robotics Research, Department of Engineering, King's College London, UK.}%
}

\begin{document}%
\maketitle%
\thispagestyle{empty}%
\pagestyle{empty}%
\renewcommand{\u}{u} 
\begin{abstract}
\Gls{LfD} is a technique that allows expert teachers to teach task-oriented skills to robotic systems. However, the most effective way of guiding \emph{novice teachers} to approach expert-level demonstrations quantitatively for specific teaching tasks remains an open question. To this end, this paper investigates the use of \MT\ to guide novice teachers to improve their teaching skills based on \emph{\RLfD}. 
The paper reports an experiment in which novices receive \MT-derived guidance to train their ability to teach a given motor skill with only $8$ demonstrations and generalise this to previously unseen ones. Results indicate that the \MT-guidance not only enhances robot learning performance by $89\%$ on the training skill but also causes a $70\%$ improvement in robot learning performance on skills not seen by subjects during training. These findings highlight the effectiveness of \MT-guidance in upskilling human teaching behaviours, ultimately improving demonstration quality in \RLfD.
\end{abstract}
\glsresetall%

\section{Introduction}%
\label{s:intro}%
\Gls{LfD} has emerged as a highly effective method to teach robots new skills by leveraging human expertise, bypassing the complexities of traditional programming, and significantly enhancing robotic deployment efficiency and reliability. However,  the efficacy and efficiency of \LfD\ heavily depend on the quality and quantity of the demonstrations, which varies substantially with the teacher's expertise\cite{correia2023survey}. This variation presents a significant challenge, as not all demonstrators are experts, and even experts cannot provide sufficient optimal demonstrations for every target skill\cite{osa2018algorithmic}.


To address these limitations, several recent studies have investigated how concepts in \emph{\MT}, a field of research concerned with devising a minimum amount of optimal data for machine learning models, can be combined with \LfD\ to train novice teachers to provide demonstrations akin to those of experts.  
For instance, \textcite{sakr2023can} demonstrated that using \AR, 
along with information entropy inspired by \MT, can reduce the number of demonstrations while maintaining teaching efficacy. \textcite{zhu2024using} proposed an \MT\ framework combined with robot \LfD\ in which trainee teachers showed a 79\% improvement in the accuracy of the taught robot motor skills. 
These studies show the potential of \MT\ to transform novices into skilled teachers who can provide high-quality demonstrations with the least demonstration quantity.

However, prior \LfD\ studies primarily focus on \emph{supervised learning problems}, limiting their applicability. In general, \RL\ is better suited than supervised learning for tasks that extend over long horizons or those that involve adaptation to changes in dynamics\cite{kumar2022should}. 
%
To this end, this paper addresses the issue of training non-expert users to teach robots through \emph{\RLfD} by employing a training process derived from \MT. It introduces a formulation of \MT\ suitable for guiding users to provide the least amount of high-quality rewards in the context of robot \RLfD, and a framework to deliver this guidance through scaffolding training. The effectiveness of the framework is evaluated through experiments in which novices are first trained to teach simple optimal control tasks where the true reward function is known. Then their overall improvement in teaching skill is measured on a new skill for which no teacher training has been given and no known rewards. Experimental results show that novices trained in this way produce demonstrations with $64\%$ higher quality on skills that do not appear in training, leading to $70\%$ improvement in learning outcomes for the robot. 
These findings suggest that \MT\ could be used to upskill workers efficiently in jobs requiring interaction with robots that learn through \RL, addressing the challenges of job displacement and the evolving needs of automation-driven industries\cite{chui2016machines,deranty2022artificial,smith2014ai,zhu_singla_zilles_rafferty}.

\begin{figure*}[t!]%
    \centering%
    \includegraphics[width=1.0\textwidth]{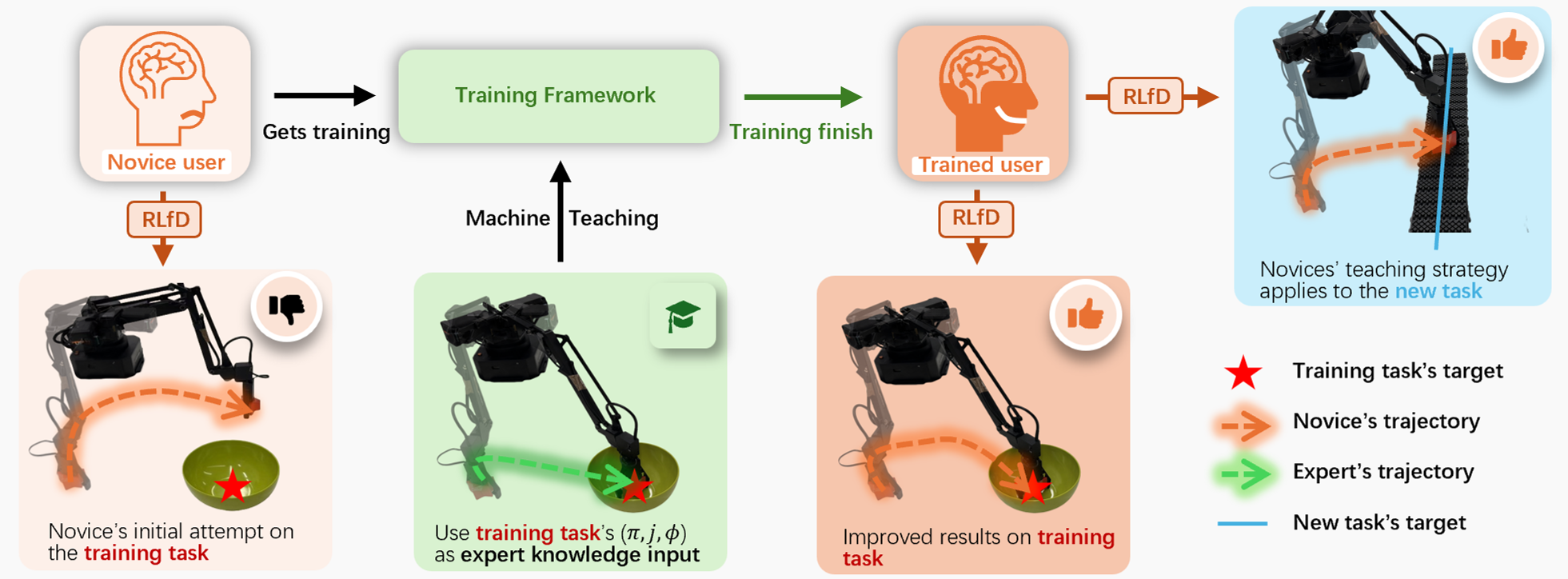}%

    \caption{Overview of training workflow. The novice user provides low-quality demonstrations, causing the robot to fail at the training task after \RLfD\ (light orange box). Using the known optimal solution to the training task (green box), \MT\ can generate guidance to be used in the training framework. Once the user has been trained, they can provide rewards more effectively, leading to improved learning outcomes for the robot both on the training task (orange box) and for new tasks not seen in training (blue box).}%
\label{f:research_framework}%
\vspace{-4ex}%
\end{figure*}%

\section{Problem Definition}%
\label{s:problem}%
\subsection{RLfD}
This paper addresses the issue of training novices to teach robot learners target skills through \RLfD. Specifically, it is assumed that target skills are optimal with respect to some reward function and can be represented in the form
\begin{align}
    \action = 
    \policy(\state) &= \argmax_{\action} \valuefunction(\state, \action) \label{e:Policyoptimise} 
\end{align}
\textbf{\vspace{-2.5ex}}
where
\begin{equation}
    \valuefunction(\state, \action)=\modelq^{\T} \featureq(\state, \action)\label{e:QbestModel}
\end{equation}
is the action-value function. Here, $\state \in \R^\dimx$ is the system state, $\action\in\R^\dimu$ is the action, $\featureq(\state,\action) \in \R^{\dimfeatureq}$ is a vector of basis functions and $\modelq\in\R^{\dimfeatureq}$ is a parameter vector corresponding to the value function for the target skill. 

The robot can be taught the target skill by providing a reward $\cost_\nd\in\R$ for the action $\action_\nd$ taken by the robot when it encounters a state $\state_\nd$, \ie $\dataset = (\state_\nd, \action_\nd)_{\nd = 1}^{\Nd}$.  The robot learns by estimating the parameters $\modelq$ of the action-value function from $\dataset$ through an \RL\ algorithm
\begin{equation}
\modelq = \A(\dataset).\label{e:RL(D)}
\end{equation}
The efficacy and efficiency of this process are significantly influenced by the quality and quantity of $\dataset$. Traditionally, reward functions are designed based on expert knowledge, however, the real-world deployment of such learners for use by non-experts, demands that means are found to \emph{train} those users to produce high-quality teaching rewards.

\subsection{Machine Teaching}
Previous studies in \emph{supervised \LfD} have shown that an effective way to train novice teachers is to find demonstrations that are close to optimal for an example task, and guide them toward using these in their teaching practice\cite{zhu2024using}. This paper adapts this approach to the \RLfD\ setting, as illustrated in \fref{research_framework}. 

In this framework, \MT\ offers a quantitative way to find the minimal amount of optimal data $\bestdataset$ for a given example skill
. It can be expressed as a bi-level optimisation\cite{zhu_singla_zilles_rafferty}
\vspace{-1ex}

\begin{align}
	\bestdataset&=\argmin_{\dataset \in \searchspaceD} 
\triskf (\modelq, \truemodelq) \label{e:MTBilevel1} \\ 
	&\mathrm{s.t.}\quad \modelq = \A(\dataset) \label{e:MTBilevel2} 
 \quad\mathrm{and}\quad \terrorf (\dataset)\le \terrorf_{max} 
\end{align}\manuallabel{e:MTBilevel3}{\ref{e:MTBilevel2}}
where $\triskf (\modelq, \truemodelq)$ is the \emph{teaching risk} and $\terrorf (\dataset)$ is the \emph{teaching effort}. The teacher's problem is to minimise the teaching risk, subject to a constraint on the effort, to derive a training set $\dataset$ from the space of possible data sets $\searchspaceD$. 
%
%
In the context of \RLfD, the \emph{learner’s} problem \eref{MTBilevel2} is to find the optimal state-action value function for a given skill, based on the demonstration $\dataset$ under a certain effort budget $\terrorf_{max}$ (e.g. least demonstrations).

Notably, an implicit assumption is that the model can only be trained via \emph{demonstrations} rather than directly giving the target parameters $\truemodelq$ to the learner. This is because in many cases, \il{\item the true reward function $\truecost$ is \emph{hard to define} or only known \emph{implicitly} and \item the \MT\ problem is computationally intractable}, particularly for complex skills where the value function lacks a closed-form solution. 
The approach proposed here improves user teaching strategies when solutions to \eref{MTBilevel1}-\eref{MTBilevel2} are available. When such solutions do not exist, it still enables effective \RLfD\ by leveraging users’ enhanced skills and their ability to generalise across \RL\ tasks.


To this end, this paper tests the following hypotheses.%
\begin{enumerate}[label=$h_\arabic*$]%
    \item \label{i:h1} When teaching through \RLfD, teaching quality is measurably improved when teachers are given guidance derived from the principles of \MT. %
    \item \label{i:h2} Teachers' improved teaching ability in one skill can be transferred to another.
\end{enumerate}%
\section{Machine Teaching Training Framework}%
\label{s:method}%
This section outlines the foundation of the \MT\ training framework used in this study. 

\subsection{\MT\ Formulation}%
The core of the proposed framework is to derive guidance from \MT\ as a training signal for novice human teachers. For this, the following modelling assumptions and design choices are made.

\subsubsection{Task Representation}
\label{s:controltask}
To balance the demands of simplicity, robustness and scalability, this paper assumes target skills can be represented in the form
\vspace{-4ex}

\begin{align}\label{e:LQR}%
    \policy&= \argmax_{\bu}{\sum_{\tk=0}^\infty\gamma^\tk\cost(\state_{\tk+1}, \action_{\tk+1})}\\
    \mathrm{s.t.}\quad& \cost(\state_{\tk}, \action_{\tk})= -\featurepi_{\tk}^\T\bQ\featurepi_{\tk}-\action_{\tk}^\T\bR\action_{\tk}\label{e:cost_LQR}\\%
    \mathrm{and} \quad&\featurepi_{\tk+1}= \bA\featurepi_{\tk}+\bB\action_{\tk}\label{e:AxBu}%
\end{align}%
\textbf{\vspace{-4ex}}
with solution
\begin{align}\label{e:LiPSkillModel} 
    \action_\tk = \target\featurepi(\state_{\tk}).
\end{align}

Here, $\state_\tk$ and $\action_\tk$ represent the state and action at the $\tk$th time step, respectively, $\featurepi(\state) \in \R^\dimfeaturepi$ is a mapping from the state to a suitable feature space (with $\featurepi_{\tk}=\featurepi(\state_\tk)$), $\bA$ and $\bB$ are matrices determining state transitions, $\gamma\in(0,1]$ is a discount factor 
and $\target \in \R^{(\dimu\dimfeaturepi) \times 1}$ is a vector containing the policy parameters. Note that, the feature space may have arbitrary complexity and may include robot-specific features (such as kinematic mappings, see \sref{evaluation}).


Within this representation, skills differ based on reward function parameters ($\bQ$\footnote{Here, the bold $\bQ$ is the state cost matrix; $\valuefunction$ is the action-value function in \RL.} and $\bR$). In \RLfD, teachers typically lack explicit knowledge of these parameters but must still teach the target skill by assigning rewards based on their understanding of the task.

\subsubsection{Modelling the Learner}
\label{s:learner}
Numerous \RL\ algorithms may be used by a robot learner to derive task-oriented policies. Here, the focus is on \RLfD\ frameworks that \il{\item are sufficiently fast to enable an interactive experience for the user, and \item do not require the user to have any expert knowledge to specify the transition dynamics}. To this end, the \RL\ algorithm used in the systems studied in this paper is \LSPI\, which is widely used in robotic control systems\cite{friedrich2019least,howard2013locally}. 

\LSPI\ works by iterating between \il{\item policy evaluation, and \item policy improvement} from an arbitrary initialisation. In its discrete-time, infinite horizon formulation, the state-action value function is defined as the expected cumulative discounted reward
\vspace{-1ex}
\begin{align}
    \valuefunction(\state, \action) &= \sum_{\tk=0}^{\infty} \gamma^{\tk} \cost(\state_{\tk+1}, \action_{\tk+1}). \label{e:Qfunction}
\end{align}

In the \emph{policy evaluation} step, \LSPI\ approximates this using a function of the form 
\begin{equation}
    \valuefunction(\state, \action)=\modelq^{\T} \featureq(\state, \action).\label{e:Qestimated}
\end{equation}
For this, it makes a closed-form estimate based on minimising the temporal difference error 
\begin{align}
    &\modelq = (\Featureq^{\T}(\Featureq - \gamma \nextFeatureq))^{-1}\Featureq^{\T}\costs
    \label{e:LSPI_solver} 
\end{align}
$\Featureq = \left( \featureq(\state_i, \action_i) \right)_{i=1}^{\numD}$, 
$\nextFeatureq = \left( \featureq(\nextstate_i, \policy(\nextstate_i)) \right)_{i=1}^{\numD}$, 
both in $\mathbb{R}^{\dimfeatureq \times \numD}$denotes the state the system transitions to when applying $\action_{\nd}$ in $\state_{\nd}$. Here, and throughout the remainder of the paper, the shorthand notation $\costs\in\R^{\Nd}$ is used to denote the rewards gathered into a column vector, \ie $\costs=(\cost_{1},\dots,\cost_{\Nd})^{\T}$.

In the \emph{policy improvement} step, the policy is estimated by substituting 
\eref{Qestimated} into \eref{Policyoptimise} and solving
\vspace{-1ex}
\begin{equation}
\frac{\partial \valuefunction}{\partial \action} = 0.\label{e:gradient_solver}
\end{equation}
Note that, this may be solved in closed form for an appropriate choice of $\featureq(\state,\action)$.

The iteration between policy evaluation and policy improvement continues until convergence is met (measured as a minimum change in the parameter $\modelq$). 

\subsubsection{Modelling the Teacher}
\label{s:teacher}
The aim of the teacher is to produce a data set $\D$ that enables the learner to estimate the desired value function and learn the target skill. Therefore, the risk function for \MT\ is
\vspace{-1ex}
\begin{equation}
    \triskf (\modelq, \truemodelq) = \|\mathbf{\modelq - \truemodelq}\|_2
\label{e:RiskFunction}
\end{equation}
 \ie the $\eltwo$-norm of the difference between $\modelq$ and $\truemodelq$.
By providing reward demonstrations reflective of the target reward function $\truecost(\state_{\tk}, \action_{\tk})$, it is expected that $\modelq$ will approach $\truemodelq$, thereby leading to better learning outcomes in terms of the policy parameters $\learnt$.

Moreover, in line with the formulation of \MT\ outlined in \sref{problem}, the teacher must teach the target skill with a fixed budget of teaching effort. In the following, the teaching budget \eref{MTBilevel3} is set as
\begin{equation} \label{e:TeachingBudget}
    \terrorf (\dataset) = \numD_{TD}
\end{equation}
where $\numD_{TD}$ is the \emph{teaching dimension} for the problem, defined as the minimum number of training items required to teach the target model to the learner. For the learner defined in \sref{learner}, the teaching dimension is $\dimfeatureq$, \ie the dimension of the features $\featureq$
\cite{JiLiu_XiaojinZhu2016}.

\subsection{\MT\ Problem}
\label{s:mt_problem}
Combining \eref{LSPI_solver}, \eref{RiskFunction} and \eref{TeachingBudget}, the optimal data for the problems considered in this paper, therefore, is
\begin{align}\label{e:Optimisation}
\bestdataset&=\argmin_{\dataset \in \searchspaceD}
\|\mathbf{\modelq-\truemodelq}\|_2 
\\
\label{e:mt2}
&\mathrm{s.t.\quad} 
\modelq = (\Featureq^{\T}(\Featureq - \gamma \nextFeatureq))^{-1}\Featureq^{\T}\costs\\
\label{e:mt3}
&\mathrm{and}\quad \terrorf (\dataset) = \dimfeatureq.
\end{align}
%

Denoting the ideal reward demonstrations as $\truecosts$, the parameter learnt by \LSPI\ from ideal teaching is
\begin{equation}
\truemodelq = (\Featureq^{\T}(\Featureq - \gamma \nextFeatureq))^{-1}\Featureq^{\T}\truecosts.
\end{equation}
Noting that the state-action pairs $(\state_\nd,\action_\nd)_{\nd=1}^{\Nd}$ are given, and noting the sub-multiplicative property of matrix norms\cite{horn2012matrix}, it can be shown that\cite{zhu2024using}
\begin{equation}\label{e:costdiff}
\|\modelq-\truemodelq\|_2 \propto\|\costs-\truecosts\|_2.
\end{equation}
Therefore, in the setting considered in this paper, training users to provide rewards that are close to the ideal rewards evaluated on the sample state action pairs is expected to lead to high-quality estimates of the value function, and in turn, improved outcomes in terms of the control policy. 

\subsection{\MT-Training Interface}
\label{s:feedback}

Training is delivered through a visual interface that allows them to explore the effect of providing different rewards on learning for a training task in a scaffolding way. This aligns with how humans naturally develop transferable skills
\cite{national2000people}. 

\begin{figure}[t!]
    \centerline{\includegraphics[width=\linewidth]{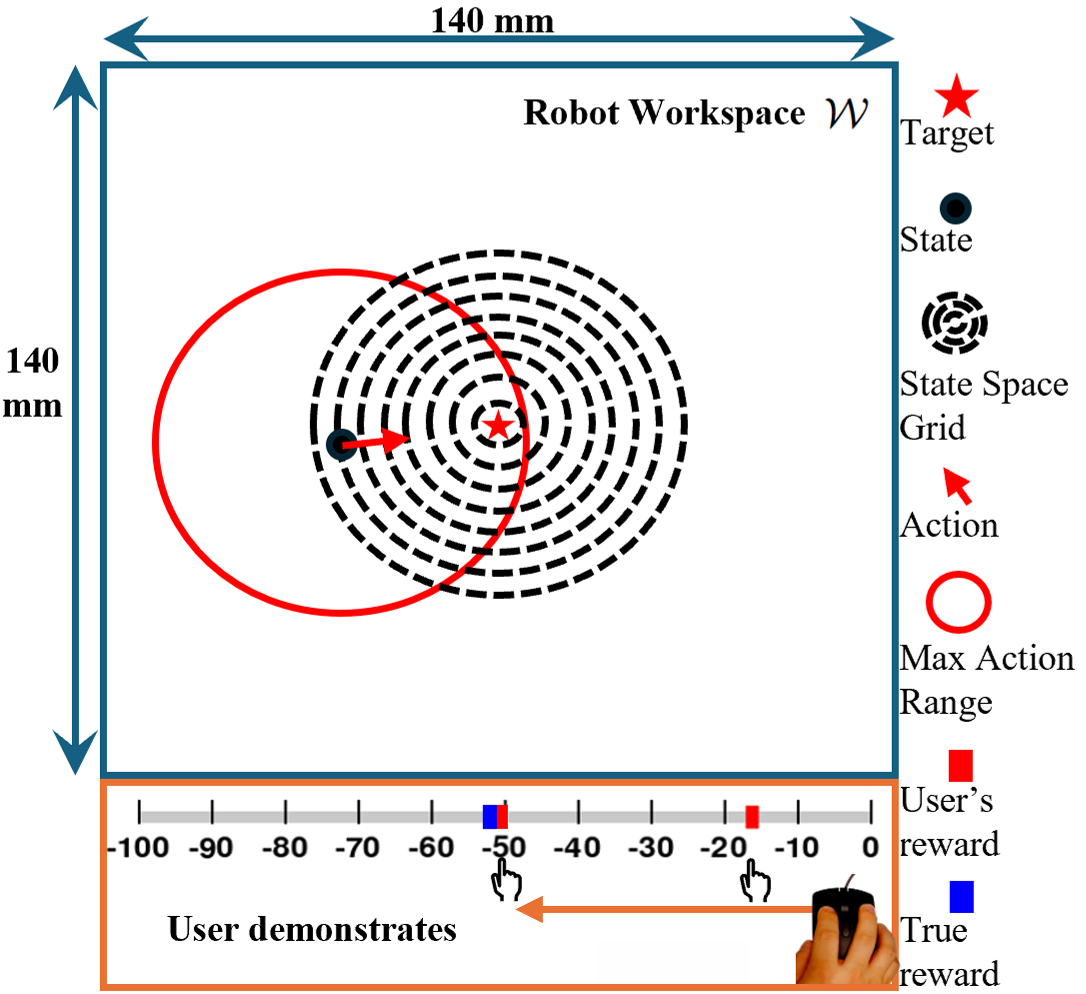}}
    \caption{Visual training interface setup for example training task \skillref{1} (see \sref{evaluation}). Trainee teachers are required to click and drag the red slider to provide reward demonstrations (orange arrow) for the key-frame state-action pair presented (black dot, red arrow). Feedback is provided in terms of the ideal reward $\truecost$ (blue bar). To aid users in calibrating the scale of rewards, the robot's workspace is demarked with a grid (black dashed lines) and a representation of the maximum admissible actions (red circle).} 
\label{f:teaching_feedback}
\end{figure}
\subsubsection{Visual Feedback  for Real-Time Reward Correction}
Research in \HCI\ has shown that visualising feedback improves learning by reducing cognitive load and facilitating quick adjustments\cite{norman1988design}. With this in mind, the interactive teaching interface used here provides state-action-reward visualisation for real-time comparison (see \fref{teaching_feedback}). 
To aid users in calibrating to the scale of the space, a grid is provided (either in polar or Cartesian coordinates) with evenly spaced intervals, with the current state visualised as a black dot. Actions $\action$ are shown as a red arrow indicating size and direction and the maximum range of actions is shown as a red circle. This clear and direct mapping reduces cognitive load, enabling users to understand and respond to feedback quickly\cite{milgram1994taxonomy}. Rewards ($\humancost$) are assigned by the user using a slider (red bar) on a linear scale that may be clicked and dragged. The slider controlled by the user remains visible throughout the demonstration. When training a user, an additional blue slider bar is shown, indicating the ideal cost ($\truecost$ as derived from \MT) to allow them to visualise the difference between the given and ideal demonstrations. 
%
%

\subsubsection{Scaffolding Training}\label{s:scaffolding}
The approach of progressively increasing task complexity, often referred to as \emph{scaffolding training} or \emph{curriculum learning}, is particularly effective in educating novices\cite{khan2011humans,bengio2009curriculum}. Here, the scaffolding training process ensures a structured progression of how reward demonstrations impact behaviour. As illustrated in \fref{real_training} and the video\footnote{The video is submitted as supplementary material.}, it consists of the following phases.
\begin{enumerate}[label=P\arabic*, ref=\arabic*, leftmargin=*]\setcounter{enumi}{2}%
\item\label{phase:3} \emph{Understanding reward magnitude consistency}: Participants learn that for same-magnitude actions directed at the target, states equidistant from the target should receive equal rewards. 
\item\label{phase:4} \emph{Identifying the action-direction-reward relationship}: Keeping state and action magnitude constant while varying action directions, participants discover how reward assignment should vary. (Counterintuitively, in the target-reaching task considered here, the reward does not depend on action directions.)
\item\label{phase:5} \emph{Identifying the state-reward relationship}: Participants learn how increasing the state (distance from the target) while keeping actions fixed affects reward values. 
\item\label{phase:6} \emph{Identifying the action-magnitude-reward relationship}: Participants learn how increasing the action magnitude while keeping the state as maximum affects reward values. Note that, the final training demo in \phaseref{5} is identical to the first step in \phaseref{6}, ensuring a smooth transition before introducing action magnitude variability (\phaseref{6}). After \phaseref{6}, participants understand that reward is the sum of independent contributions from state and action magnitudes. (Counterintuitively, the 8th demonstration in \phaseref{6} reaches the target with the largest action but receives the lowest ($-100$) reward. This is because the reward discourages large actions, ensuring a smoother trajectory and preventing potential damage to the robot)
\item\label{phase:7} \emph{Developing robust reward assignment}: To enhance their ability to assign rewards without predefined rules, participants practice inferring rewards for randomly selected state-action pairs, strengthening their implicit understanding before encountering novel tasks.
\end{enumerate}
Note that, the generic nature of this curriculum makes it easily adaptable to training reward assignments for a wide variety of \RLfD\ tasks. 
\vspace{-1ex}
\section{Experiments} 
\label{s:evaluation}
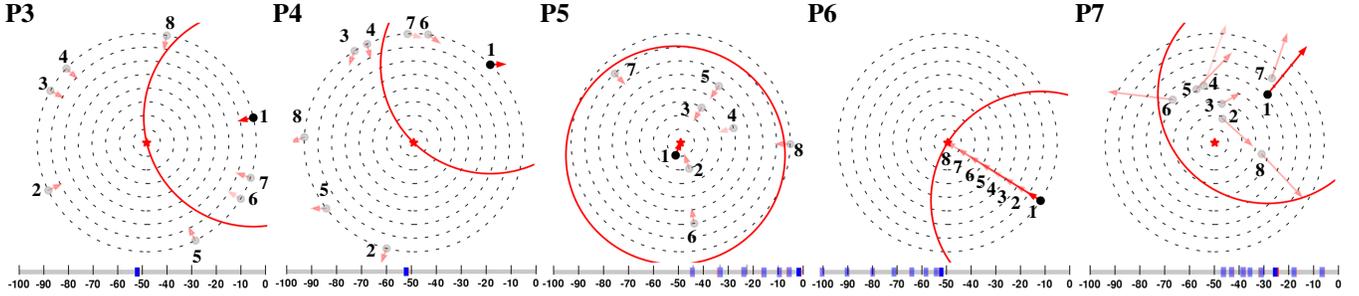
\begin{figure*}[ht!]
    \centering
    \begin{overpic}[width=1.0\textwidth]{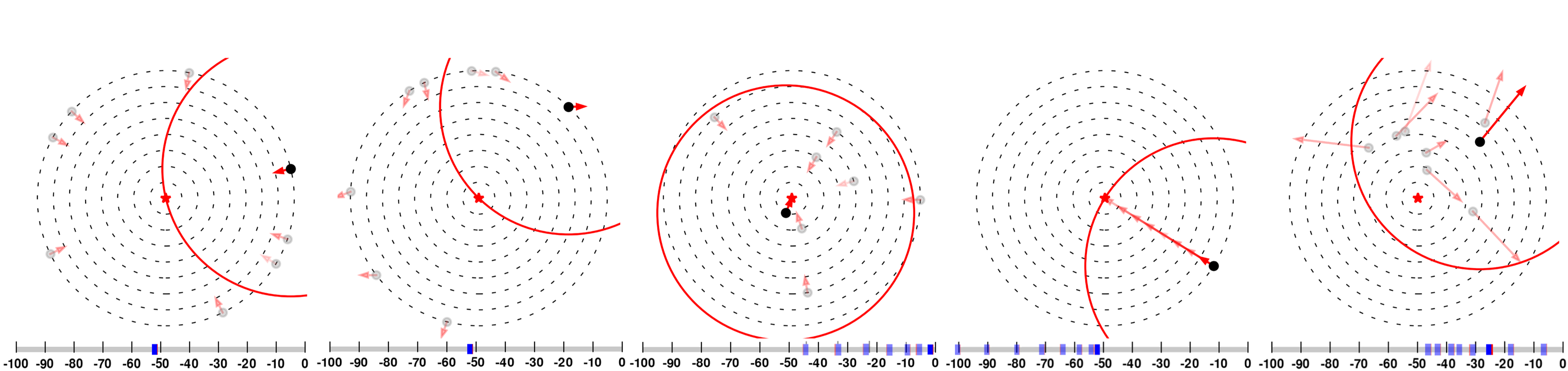}
        \put(0,20){\textbf{\phaseref{3}}} 
        \put(20,20){\textbf{\phaseref{4}}}
        \put(40,20){\textbf{\phaseref{5}}}
        \put(60,20){\textbf{\phaseref{6}}}
        \put(80,20){\textbf{\phaseref{7}}}
        \put(19, 12.5){\scriptsize$\textbf{1}$}
        \put(2, 7){\scriptsize$\textbf{2}$}
        \put(2.5, 15){\scriptsize$\textbf{3}$}
        \put(4, 17){\scriptsize$\textbf{4}$}
        \put(14, 2){\scriptsize$\textbf{5}$}
        \put(18.2, 6.3){\scriptsize$\textbf{6}$}
        \put(19, 7.5){\scriptsize$\textbf{7}$}
        \put(12, 19.5){\scriptsize$\textbf{8}$}
        
        \put(36, 17.5){\scriptsize$\textbf{1}$}
        \put(27, 2.5){\scriptsize$\textbf{2}$}
        \put(25, 18.5){\scriptsize$\textbf{3}$}
        \put(27, 19){\scriptsize$\textbf{4}$}
        \put(23.5, 7){\scriptsize$\textbf{5}$}
        \put(31, 19.5){\scriptsize$\textbf{6}$}
        \put(30, 19.5){\scriptsize$\textbf{7}$}
        \put(21.5, 12.5){\scriptsize$\textbf{8}$}
        
        \put(49, 9.5){\scriptsize$\textbf{1}$}
        \put(51.5, 8.5){\scriptsize$\textbf{2}$}
        \put(50.5, 13){\scriptsize$\textbf{3}$}
        \put(54, 12.6){\scriptsize$\textbf{4}$}
        \put(52, 15.5){\scriptsize$\textbf{5}$}
        \put(46.5, 16){\scriptsize$\textbf{7}$}
        \put(51, 3.5){\scriptsize$\textbf{6}$}
        \put(59, 10){\scriptsize$\textbf{8}$}

        \put(76.5, 5.4){\scriptsize$\textbf{1}$}
        \put(75.2, 6){\scriptsize$\textbf{2}$}
        \put(74.2, 6.5){\scriptsize$\textbf{3}$}
        \put(73.4, 7.1){\scriptsize$\textbf{4}$}
        \put(72.6, 7.6){\scriptsize$\textbf{5}$}
        \put(71.8, 8.1){\scriptsize$\textbf{6}$}
        \put(70.9, 8.7){\scriptsize$\textbf{7}$}
        \put(70, 9.3){\scriptsize$\textbf{8}$}     
        
        \put(94, 13){\scriptsize$\textbf{1}$}
        \put(91.5, 12.5){\scriptsize$\textbf{2}$}
        \put(89.7, 13.3){\scriptsize$\textbf{3}$}
        \put(90, 15){\scriptsize$\textbf{4}$}
        \put(88, 14.5){\scriptsize$\textbf{5}$}
        \put(86.5, 13){\scriptsize$\textbf{6}$}
        \put(93.5, 15.7){\scriptsize$\textbf{7}$}
        \put(93.5, 8.5){\scriptsize$\textbf{8}$}

    \end{overpic}
    
    \caption{Scaffolding training (\phaseref{3}–\phaseref{7}) for \skillref{1} (see \sref{evaluation}). The ideal reward for the illustrated state (black dots)-action (red arrows) pairs is shown in blue on the slider at the bottom. Example state-action-reward pairs are shown in sequence, with the bold one representing the 1st demonstration, while the faded ones represent future demonstrations.}
    
    \label{f:real_training}
    \vspace{-4ex}
\end{figure*}
In this section, the efficacy of the proposed framework 
is evaluated 
with the aim of testing hypotheses \ref{i:h1}-\ref{i:h2} (see \sref{problem}). 
\vspace{-4.5ex}
\subsection{Setup}
\label{s:setup}
\vspace{-0.5ex}
In this study, subjects are tasked with teaching a two-link, kinematically-controlled robot arm (uArm Swift Pro, UFactory, China) and a class of \RL\ skills involving reaching a given target in the end-effector workspace. 
This encompasses reaching target points at different locations, and target lines with different orientations. The state feature vector is $\featurepi(\state)=(\statei,\stateii)^\T$ where $\statei$ and $\stateii$ represent the 2D location of the end-effector with $-70 \leq \r_i \leq 70 \SI{}{\milli\meter}, i\in\{1,2\}$. The actions are $\action=(\ui, \uii)^\T$ the desired change in the end-effector position within one time-step. The system operates at sample rate $100\, Hz$. 
Visual feedback is displayed on Dell U2415 with a $1920\times1200$ display resolution.

The skills that must be taught to the robot are:
\begin{enumerate}[label=S\arabic*, ref=\arabic*] 
	\item\label{skill:1} \emph{Reaching to a point $\desired{\br}$ from any start state}, and 
    \item\label{skill:2} \emph{Reaching to a line $\cos{(\alpha)}\desired{\statei}+\sin{(\alpha)}\desired{\stateii}+d=0$ from any start state}, where $\alpha$ is the rotation angle and $d$ is the offset. 
\end{enumerate} 
Note that, such skills can be represented through \eref{cost_LQR} where 
\vspace{-1.5ex}
\begin{equation}%
    \bR=\beta\I,%
\end{equation}%
\vspace{-2.5ex}
with $\bQ=\bR$ for \skillref{1} and
\begin{align}
    \bQ&=\beta\rotation^\T\begin{pmatrix}
\epsilon & 0 \\
0 & 1
\end{pmatrix}\rotation,\label{e:cost_S2_Q}
\end{align}

for \skillref{2}. Here, $\I$ is the identity matrix, $\beta=0.01$ and $\rotation$ is a rotation matrix. Note that, $\epsilon=
10^{-15}$ is a regularisation parameter used to avoid numerical instability in solving the optimal control problem. In the below, $\desired{\br}=(0,0)^\T$, $d=0$ and $\alpha=\pi\,\SI{}{\radian}$.

To teach these skills, the subject must provide rewards $\humancosts$ 
for a pre-specified set of state-action tuples $(\state_\nd,\action_\nd)_{\nd=1}^{\Nd}$ to the robot learner. The learner uses \LSPI\ with $\featureq(\state, \action)=(\statei^2, \stateii^2, \ui^2, \uii^2, \statei\ui, \statei\uii, \stateii\ui, \stateii\uii)$, therefore $\Nd=\exptiNd$ is the minimum number of data points required to solve \eref{LSPI_solver} and forms the 
teaching budget $\terrorf (\dataset)$ for this problem. 

\subsection{Protocol}
\label{s:experiment_procedures}

\vspace{-0.5ex}
The experiment is a between-subjects study designed to evaluate the effect of training on participants' ability to teach motor skills through \RLfD. From a power analysis (HHU G*Power), ten participants per group (total $n_{p}=20$) are sufficient for detecting the effect size with a 0.1 error probability and 0.8 power. Participants are randomly assigned to either the control (non-guided) or target (guided) group and are instructed on how to use the teaching interface. The experiment then proceeds through the following phases:
\begin{enumerate}[label=P\arabic*\hspace{0.85em}, ref=\arabic*, leftmargin=*]
    \item\label{phase:1} \textit{Skill 1, Test 1:} Participants provide reward demonstrations $(\humancost_\nd)_{\nd=1}^8$ to teach \skillref{1} \emph{without guidance}. In the experiments reported here, the state-action tuples are sampled uniform-randomly, \ie $\r_{i,\nd}\sim\U[-35,35]\text{mm}$ and $\delta\r_{i,\nd}\sim\U[-35,35]\text{mm}$ for $i\in\{1,2\}$ and $1\le\nd\le\Nd$.
    \item\label{phase:2} \textit{Skill 2, Test 1:} Participants repeat \phaseref{1}, but provide reward demonstrations to teach for \skillref{2}.
    \item[\phaseref{3}-\ref{phase:7}]\label{phase:_train} \textit{Scaffolding Training Phases:} Participants teach \skillref{1} throughout the scaffolding training scenarios described in \sref{scaffolding} and \fref{real_training}. In these phases, guidance is provided to the target group, but not to the control group.%
    \setcounter{enumi}{7}%
    \item\label{phase:8} \textit{Skill 2, Test 2:} Participants repeat \phaseref{2}.
    \item\label{phase:9} \textit{Skill 1, Test 2:} Participants repeat \phaseref{1}.
\end{enumerate}
Note that, with this experimental design, hypothesis \ref{i:h1} (improvement) is tested by comparing teaching performance in \phaseref{1} and \phaseref{9}, and \ref{i:h2} (transfer) is tested by comparing \phaseref{2} and \phaseref{8}. Here, teaching performance is measured through the following metrics:
\begin{enumerate}[label=$m_\arabic*$, ref=$m_\arabic*$, leftmargin=*]
    \item\label{i:ade} The \emph{\gls{ADE}} 
    \textbf{\vspace{-1.5ex}}space
    $$\Eade = \sum_{\nd=0}^{\Nd}\|\humancost_{\nd} - \truecost_{\nd}\|.$$
    This quantifies the difference between optimal and teacher-provided demonstrations.
    \item\label{i:rmse} The \emph{\gls{ARMSE}} 
    \textbf{\vspace{-1ex}}
    $$\armse = \frac{1}{M}\sum_{m=1}^{M}\sqrt{\frac{1}{\Tk} \sum_{\tk=0}^{\Tk-1} \|\learntstate_{\tk} - \truestate_{\tk} \|^2}$$
    This measures the average difference in trajectories generated by the policy taught with the ideal rewards $\truecosts$ against those from the policy taught with the teacher-provided costs $\humancosts$ from different start states. 
    \item\label{i:tc} The \emph{\gls{ATC}} accumulated by the aforementioned trajectories with respect to the true reward function, \ie 
    \textbf{\vspace{-1ex}}
    $$\atc = \frac{1}{M}\sum_{m=1}^{M}\sum_{\tk=0}^{\Tk-1} \truecost(\state_{\tk}, \action_{\tk}).$$ 
    \textbf{\vspace{-2ex}}
\end{enumerate}

In the results reported below, \ref{i:rmse} and \ref{i:tc} are 
computed on $M = 100$ trajectories with a duration of $1$ second (\ie $\Tk=100$) with start states drawn uniform-randomly within the workspace of the robot.


%
%
%
%
%

\subsection{Results}
\vspace{-1ex}

\begin{figure}[t!]
    \centering
    \begin{overpic}[width=0.49\textwidth]{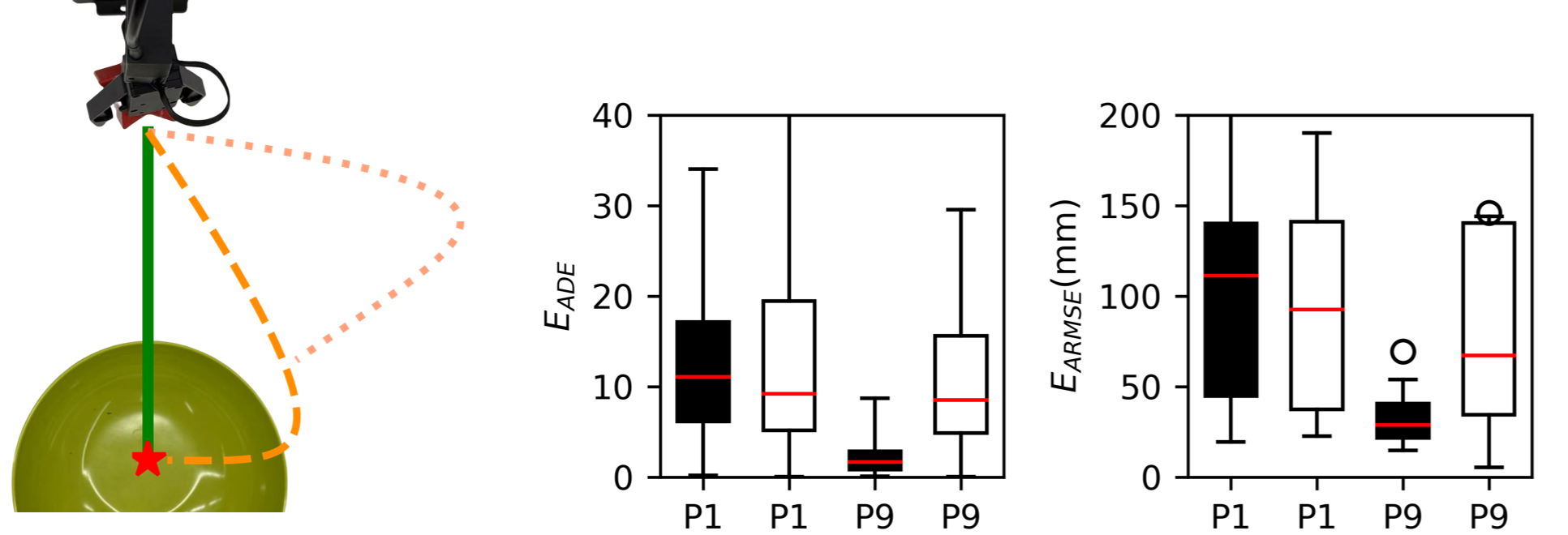}
        \put(19,29){\ref{f:mt_result_1a}}
        \put(50,29){(b)}
        \put(83,29){(c)} 
    \end{overpic}
    \caption[singlelinecheck=off]{Comparison of subjects' performance in teaching \skillref{1}: \cl{\item\label{f:mt_result_1a} Top view of a trajectory taught by a representative subject from the target group in \phaseref{1} (dotted orange line) versus \phaseref{9} (dashed orange line), where the gripper is reaching to place an object at the target in the bowl. \item $\Eade$ and \item $\armse$ in \phaseref{1} and \phaseref{9}, for the target (black box) and control (white box) groups. The red lines are medians. Black circles indicate outliers (defined as those lying outside the $0$-$90$ percentile range).}}
    \label{f:mt_result_1}
\end{figure}

\begin{figure}[t!]
    \centering
        \begin{overpic}[width=0.49\textwidth]{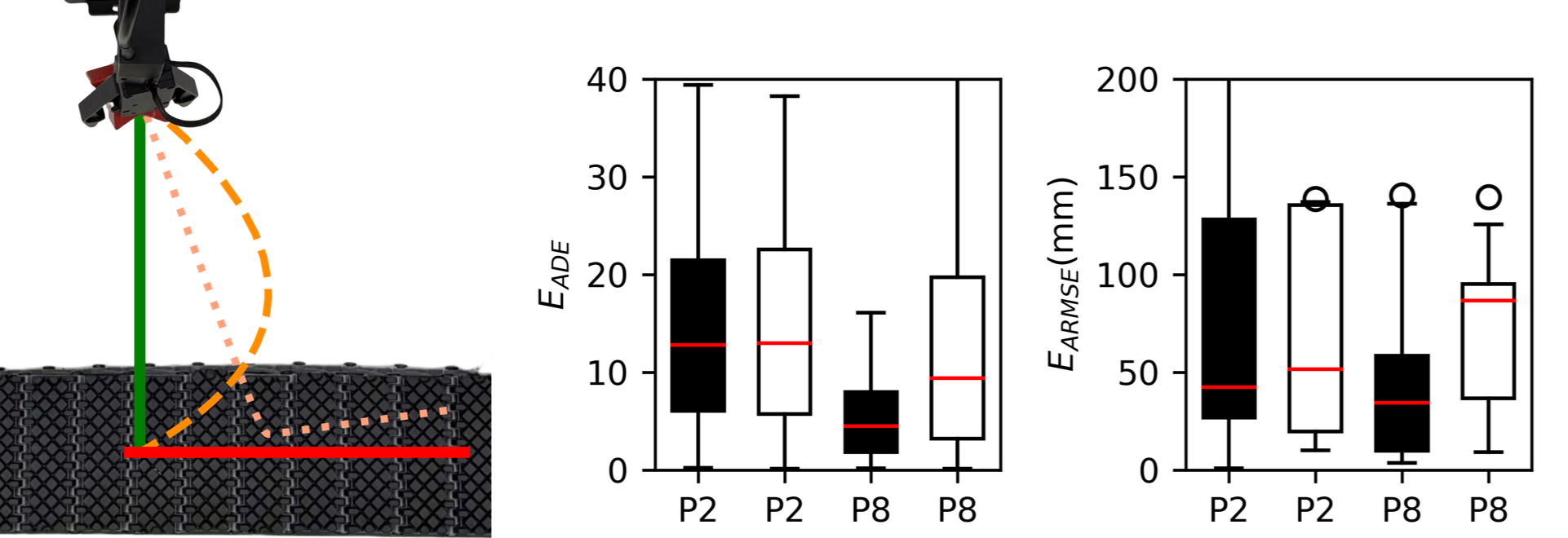}
        \put(19,31){\ref{f:mt_resuly_2a}}
        \put(50,31){\ref{f:mt_resuly_2b}}
        \put(83,31){\ref{f:mt_resuly_2c}} 
    \end{overpic}
    \caption[singlelinecheck=off]{Comparison of subjects' performance in teaching \skillref{2}: \cl{\item\label{f:mt_resuly_2a} Top view of a trajectory taught by a representative subject from the target group in \phaseref{2} (dotted orange line) versus \phaseref{8} (dashed orange line), where the gripper is reaching to place an object onto the conveyor belt (red line). \item\label{f:mt_resuly_2b} $\Eade$,  and \item\label{f:mt_resuly_2c} $\armse$ in \phaseref{2} and \phaseref{8}, for the target (black box) and control (white box) groups. Black circles indicate outliers (defined as those lying outside the $0$-$90$ percentile range).}}%
    \label{f:mt_result_2}
\end{figure}

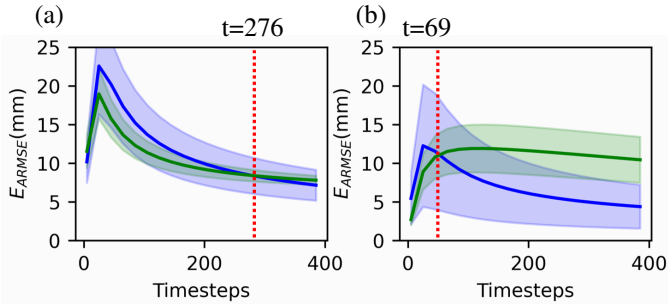
\begin{figure}[t!]

    \centering

    \begin{overpic}[width=0.5\textwidth]{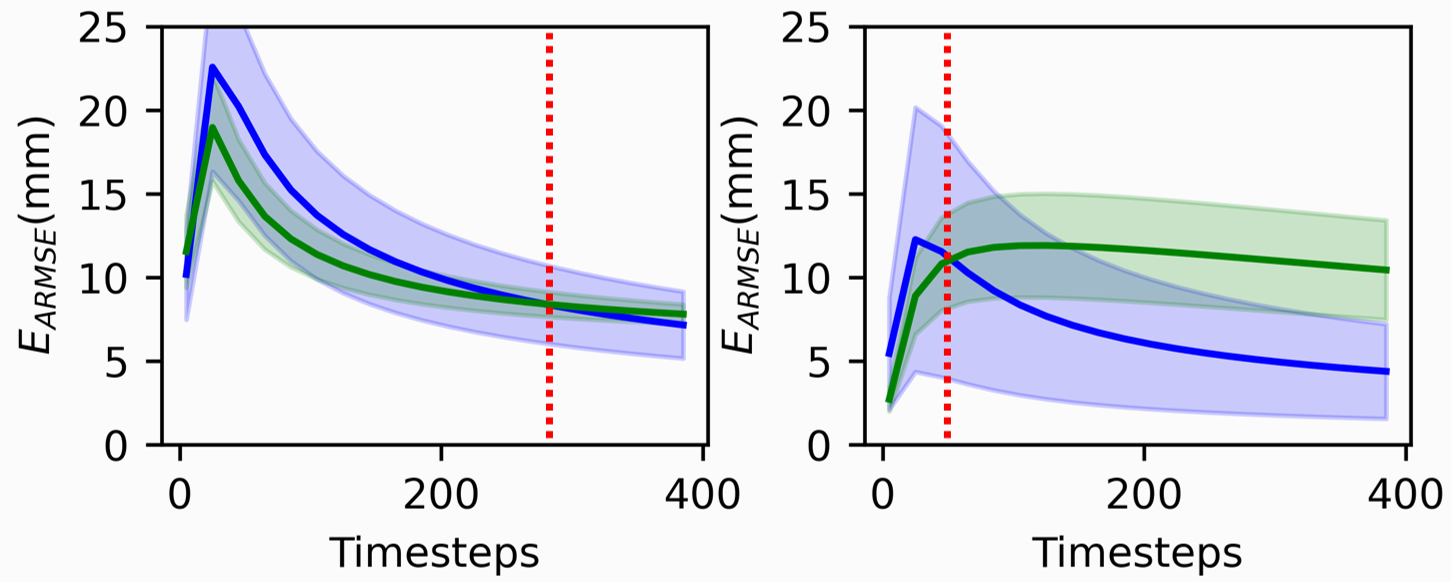}
        \put(5, 42){(a)}
        \put(53, 42){(b)}
        \put(33,40){t=276}
        \put(60,40){t=69}
    \end{overpic}
    \textbf{\vspace{-3ex}}
    \caption[singlelinecheck=off]{Comparison of learning outcomes for teachers trained in supervised \LfD\ (green) versus \RLfD\ (blue) for when using policies over different time horizons. Shown is the mean$\pm$s.d. \ARMSE\ of the learnt policy for \cl{\item \skillref{1} and \item\skillref{2}}.
    }
    \label{f:mt_result_3}
\end{figure}

\subsubsection*{Effectiveness of \MT-based training (\ref{i:h1})} The target group shows an $83\%$ reduction in $\Eade$ (Wilcoxon Test, $p = 9.78\times10^{-14}$), an $89\%$ reduction in $\armse$ ($p = 0.013$), and a $98\%$ reduction in $\atc$ ($p = 0.006$). These reductions confirm that guidance significantly improves teaching and learning outcomes in \skillref{1}. The control group showed no significant changes ($p > 0.1$ in all tests), indicating the absence of training effects without guidance. These results robustly support \ref{i:h1}, demonstrating that teaching performance is enhanced on the skills used for training. An illustration of the learning outcomes for the robot when taught by a representative subject in the target group is shown in \fref{mt_result_1}\ref{f:mt_result_1a}. As can be seen, the trajectory generated pre-training fails to reach the target, whereas that generated post-training successfully hits it. 
\subsubsection*{Transfer of Teaching Ability (\ref{i:h2})}
For skill \skillref{2}, the target group achieves a $64\%$ reduction in $\Eade$ ($p = 4.73\times10^{-9}$), indicating improved demonstration quality is transferred to this skill. Changes in $\armse$ ($p = 0.27$, $70\%$ reduction) and $\atc$ ($p = 0.16$, $91\%$ reduction) are not statistically significant. This discrepancy between demonstration quality and robot performance may be attributed to reward ambiguity: in this task multiple policies yield the same cumulative rewards, reducing the sensitivity of $\armse$ and $\atc$ as measures of teaching performance. The control group shows no significant improvement in teaching performance or learning outcomes ($p>0.05$ in all tests). The results are illustrated in \fref{mt_result_2}.
\subsubsection*{Supervised \LfD\ versus \RLfD}
To further evaluate the proposed framework the experiment is repeated to compare the robot learning outcomes achieved by those trained to teach with \RLfD\ against those trained in supervised \LfD. To this end, a further cohort of $n_{p}=10$ participants are trained to teach \skillref{1} using the \MT-protocol described by \textcite{zhu2024using}, whereby subjects directly provide demonstrations of the required actions on the corresponding states, and the robot learns through supervised ridge regression. The resultant policies are then used to generate trajectories for \skillref{1} and \skillref{2} of different duration ($1\le\tk\le400$). The results are summarised in \fref{mt_result_3}. As can be seen, for short duration ($\tk\le279$ in \skillref{1} and $\tk\le69$ in \skillref{2}), the \ARMSE\ of the robot's policy is lower when taught by trained teachers using supervised \LfD, whereas for longer duration, trained teachers using \RLfD\ achieve better outcomes in terms of the long-term stability of the learnt behaviour.

\section{CONCLUSIONS}
\label{s:conclusion}
This paper proposes a scaffolding, \MT-based framework to train novices to teach robots skills through \RLfD. The findings demonstrate that guidance significantly improves demonstration quality in the skills used for training the teacher and, critically, causes enhanced teaching quality in new, unseen tasks for which training has not been received. Improvements to robots' learning outcomes are particularly pronounced in long-horizon tasks. 
This may have far-ranging impact since it demonstrates that novice teachers can be upskilled very quickly without engaging in high-level, long-term education programmes. In future work, the framework will be extended to the teaching of more complex tasks, for instance, by using local optimal control methods\cite{tedrake2009lqr}.


\addtolength{\textheight}{0cm}   

{\scriptsize\printbibliography}%

\end{document}